\documentclass[twocolumn]{article}

\usepackage{microtype}
\usepackage{graphicx}
\usepackage{subcaption}
\usepackage{booktabs}
\usepackage{hyperref}

\usepackage[linesnumbered,ruled,vlined]{algorithm2e}
\usepackage{geometry}

\usepackage{geometry}
\usepackage{caption}
\usepackage{amsmath}
\usepackage{amssymb}
\usepackage{mathtools}
\usepackage{amsthm}
\usepackage{amsfonts}
\usepackage{bm}
\usepackage{tabularx}
\usepackage{float}
\usepackage{multirow}
\usepackage{colortbl}
\usepackage{longtable}
\usepackage[capitalize,noabbrev]{cleveref}
\usepackage[textsize=tiny]{todonotes}
\usepackage{adjustbox}
\usepackage[numbers,sort&compress]{natbib}
\usepackage{dblfloatfix}
\usepackage{sectsty}
\usepackage{ragged2e}

\newif\ifdraft

\drafttrue

\ifdraft
 \newcommand{\PF}[1]{{\color{red}{\bf PF: #1}}}
 
  \newcommand{\AF}[1]{{\color{blue}{\bf AF: #1}}}
 
 \newcommand{\HL}[1]{{\color{orange}{\bf HL: #1}}}

\newcommand{\CD}[1]{{\color{purple}{\bf CD: #1}}}

\newcommand{\NT}[1]{{\color{teal}{\bf NT: #1}}}

\newcommand{\MX}[1]{{\color{cyan}{\bf MX: #1}}}

 \newcommand{\YXY}[1]{{\color{pink}{\bf YXY: #1}}}
 
\else
 \newcommand{\PF}[1]{}
 
 \newcommand{\AF}[1]{}
 
 \newcommand{\HL}[1]{}
 
 \newcommand{\CD}[1]{}
 
 \newcommand{\NT}[1]{}
 
 \newcommand{\MX}[1]{}
 
 \newcommand{\YXY}[1]{}
 
\fi

\newcommand{\ours}{{\it ACG}}

\newcommand{\State}{\mathbf{x}}

\newcommand{\Mask}{\mathcal{M}}

\definecolor{htblue}{rgb}{0.0, 0.0, 0.55}

\ifdraft
  \newcommand{\HT}[1]{{\color{htblue}{\bf HT: #1}}}
\else
  \newcommand{\HT}[1]{}
\fi
\renewcommand{\arraystretch}{1.2}
\newcommand{\alglinehl}[2]{\colorbox{#1}{\strut #2}}

\theoremstyle{plain}

\theoremstyle{definition}

\theoremstyle{remark}

\title{Annealed Co-Generation: Disentangling Variables \\ via Progressive Pairwise Modeling}

\author{%
  Hantao Zhang\thanks{Equal contribution.}\quad
  Jieke Wu\footnotemark[1]\quad
  Mingda Xu\quad
  Xiao Hu\quad
  Yingxuan You\quad
  Pascal Fua\\[0.5em] 
  {\normalfont\small CVLab, EPFL}
}
\date{}

\begin{document}
\maketitle
\begin{abstract}

For multivariate co-generation in scientific applications, we advocate pairwise block rather than joint modeling of all variables. This design mitigate the computational burden and data imbalance. To this end, we propose an \textbf{Annealed Co-Generation (ACG)} framework that replaces high-dimensional diffusion modeling with a low-dimensional diffusion model, which enables multivariate co-generation by composing pairwise variable generations. We first train an unconditional diffusion model over causal variables that are disentangled into pairs. At inference time, we recover the joint distribution by coupling these pairwise models through shared common variables, enabling coherent multivariate generation without any additional training. By employing a three-stage annealing process—Consensus, Heating, and Cooling—our method enforces consistency across shared common variables and progressively constrains each pairwise data distribution to lie on a learnable manifold, while maintaining high likelihood within each pair. We demonstrate the framework's flexibility and efficacy on two distinct scientific tasks: flow-field completion and antibody generation. All datasets and code will be made publicly available upon publication.
\end{abstract}
\section{Introduction}

\begin{figure}[ht]
    \centering
    \includegraphics[width=0.9\linewidth]{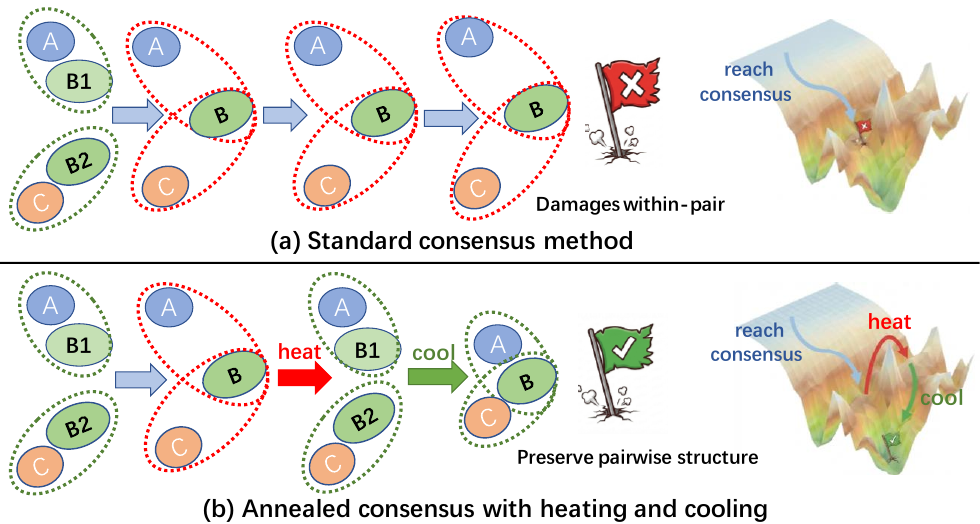}
    \caption{\textbf{Naive consensus vs. annealed consensus}. (a) Given two diffusion processes that generate variable pairs $(A,B_1)$ and $(B_2,C)$, if we wish t$B_1$ and $B_2$  to be the same $B$ in the end, we can enforce consensus at each step of the process.  However, this can easily yield implausible (low-likelihood) pairwise samples.
 (b) To avoid this, \ours{} introduces a heating--cooling schedule that temporarily splits \(B\) into \(B_1\) and \(B_2\), enabling the model to preserve strong within-pair dependencies while refining the solution through multiple re-generations that help it escape local minima before restoring agreement.}
     \label{fig:compare}
\end{figure}

Probabilistic generative models such as diffusion~\cite{Song21c} and Flow Matching~\cite{Lipman22} have revolutionized high-dimensional data synthesis for scientific discovery, enabling applications such as generating coherent 3D structures and functional biological molecules. They typically start from a sample randomly drawn from a standard normal distribution and progressively transform it into a sample from a target distribution. These methods are remarkably effective when modeling a single variable.

However, a more complex challenge arises when we aim to jointly generate multiple variables while accounting for the correlations among the resulting samples. For instance, in antigen–antibody co-design for drug discovery, preclinical animal studies are required before clinical trials~\cite{watson2023rfdiffusion,wang2025iggm,boltzgen2025}. As they are often done in mice, there is an incentive for generating antibodies that are compatible with both murine and human versions of the target antigen. Yet, despite recent progress in off-the-shelf foundation models such as BoltzGen~\cite{boltzgen2025}, current approaches are often limited to generating only a single antigen–antibody pair at a time, rather than supporting flexible multi-variable co-generation. Concretely, existing schemes can generate antigen/antibody pairs ($A$, $B_1$) for mice and similar pairs  ($A$, $B_2$) for humans. However, if we wish $B_1$ and $B_2$ to be the same, we must enforce it. A similar situation obtains in aerodynamics because complex geometries and unforeseen disturbances can degrade flow-field measurements.  A similar issue arises in aerodynamics: complex geometries and disturbances can corrupt flow-field measurements, leaving missing regions~\cite{luo2024deep}. Instead of conditioning on the entire field~\cite{shu2023physics}, we split it into patches and consider a missing segment \(B\) with upstream \(A\) and downstream \(C\) (left-to-right flow). Inpainting from the left yields \((A,B_1)\), while inpainting from the right yields \((B_2,C)\); a coherent reconstruction requires \(B_1 = B_2\).

To address this, some works adopt graph-based formulations~\cite{chamberlain2021grand}. Unfortunately, such methods typically assume or learn a fixed graph structure during training, which limits their ability to adapt the dependency structure at inference time---often a critical requirement in AI for Science, where one would like to reuse existing foundation models without retraining. This limitation is particularly restrictive in scientific settings where the data are high-dimensional and fully observed joint samples are scarce, making it impractical to retrain large models for every new dependency pattern. Moreover, when variable representations are high-dimensional but the underlying graph topology is simple (e.g., essentially tree-structured), graph-based models may still struggle to capture rich interactions within and across variable subsets. A flexible and efficient method is therefore needed to exploit the well-defined graphical structure that often exists between subsets of variables at inference time, enabling scalable composition beyond pairwise generation. To mitigate this issue, another line of work models pairwise relationships~\cite{zhang2023diffcollage}, as illustrated in Fig.~\ref{fig:compare}(a). However, enforcing global consistency through the shared intersection variable $B$ is challenging. In particular, naive pairwise composition can easily lead to low-likelihood samples for each pairwise factor and the resulting inference procedure can become unstable and get trapped in poor local minima.

Thus, we need a method that not only reaches consensus on shared variables when composing different pairs, but also preserves intrinsic within-pair relationships. Inspired by simulated annealing~\cite{Kirkpatrick83}, we propose Annealed Co-Generation (\ours{}), a generic framework for multi-variable joint generation. \ours{} enforces shared-variable consistency and within-pair relational constraints through a dynamic annealing search. We rely on diffusion schemes to produce variable pairs. But instead of forcing agreement on shared variables at every diffusion step as depicted by Fig.~\ref{fig:compare}(a),
\ours{} iteratively alternates among three phases---\textit{consensus} (linking different pairs through their shared variables), \textit{heating} (rolling back the generation process by re-noising), and \textit{cooling} (generating new pairs from starting from the re-noised samples)---as illustrated by Fig.~\ref{fig:compare}(b). We demonstrate \ours{} on two distinct and substantially different tasks—multi-antigen protein generation and flow-field inpainting—and achieve improved efficiency and accuracy, along with greater flexibility in satisfying multiple constraints. Specifically, in protein design, \ours{} outperforms approaches that design proteins against each antigen independently as well as other consensus-style baselines; in flow-field inpainting, it improves upon methods that condition on the entire field.

In short, our contribution is a method that can sample the full joint distribution of multiple variables by only relying on existing pairwise foundations, which reduces effective dimensionality and enables flexible causal interventions at test time. It captures complex within-pair causal dependencies via annealing rather than previous consensus-style methods.

\section{Related Work}

\subsection{Causality for generative AI for Science}
Across scientific domains, researchers have increasingly explored generative models as tools for capturing and leveraging domain causal structure, including tasks such as molecular structure discovery and generation~\cite{chamberlain2021grand} and flow-field synthesis~\cite{shehata2025improved,li2024synthetic}. A line of work attempts to formalize diffusion models from a graphical (causal graph) perspective~\cite{chamberlain2021grand,thorpe2022grand++}. While this perspective is well-suited to specific domains such as molecular discovery, its limitations become more apparent in broader AI-for-Science settings. First, many scientific applications do not require highly complex causal mechanisms: in scenarios such as flow fields
~\cite{fukami2023super}and antibody design\cite{boltzgen2025,martinkus2023abdiffuser}, variables often exhibit only simple dependency patterns (frequently well approximated by a tree), which do not naturally warrant a complex graph structure. Moreover, we would like the causal structure among variables to be constructed automatically at test time, rather than being pre-specified during training~\cite{vignacdigress,wangdiffusion}. Second, unlike molecular design, where the nodes in the graph can be relatively lightweight, the variables that participate in causal relationships in many AI-for-Science problems are often \emph{heavy}—i.e., they live in complex structured spaces and incur substantial representation and storage costs. Prior generative approaches for AI for Science have not fully addressed this requirement. To address this problem, we advocate training pairwise block models rather than jointly modeling all variables. At inference time, we couple these pairwise models through shared variables to recover the joint distribution, enabling coherent multivariate generation without any additional training.

\subsection{ Generating Samples from Multiple Distributions}

Some prior work has attempted to divide a joint diffusion process into multiple parts and then study the causal structure among them. For example, Disentangled Representation \cite{song2023flow} aims to disentangle the processes responsible for different types of feature variation; however, it mainly targets image style transformation and does not explicitly model interactions among distinct variables. In contrast, Gist \cite{weilbach2023graphically} explicitly defines dependencies (i.e., causal relationships) among internal variables via a graph, and enforces the diffusion model’s denoising dynamics to follow this graph structure. A key limitation is that, for high-dimensional data, the model must learn all $N^2$ pairwise relations. Moreover, once training is finished, the implied causal structure is effectively fixed, making it difficult to flexibly modify or reconfigure the causal relationships at test time.

The problem of generating samples that satisfy multiple simultaneous constraints has been addressed through several paradigms. Composable Diffusion~\cite{liu2022compositional} enables the combination of multiple conditional diffusion models by composing their noise predictions, but lacks a principled mechanism to handle temporal synchronization or resolve conflicts between constraints. StructureDiffusion \cite{liu2023structure} extends this to structured data but still relies on simple arithmetic combinations. Concurrent work on multi-conditioning \cite{liu2023multicond} explores product-of-experts formulations, but these methods operate at a fixed timestep resolution and do not account for the temporal evolution of constraint satisfaction.

Beyond point-wise composition, several works have explored the synthesis of large-scale content through the fusion of local diffusion paths. MultiDiffusion \cite{bar-tal2023multidiffusion} provides a training-free framework for high-resolution image generation and panorama synthesis by optimizing for a global image that is consistent across overlapping local windows. Similarly, DiffCollage \cite{zhang2023diffcollage} utilizes a factor graph representation to parallelize the generation of large-scale content while maintaining boundary consistency. These approaches share a common goal with our multi-target generation framework: achieving a consensus among multiple (spatial) constraints. However, while they focus on spatial consistency in pixel space, our Progressive Coupling mechanism generalizes this idea to arbitrary vector spaces and explicitly models the phase transition between independent exploration and collective commitment.

\iffalse

\fi

\section{Methodology}
\label{sec:method}
We propose \textbf{Annealed Co-Generation (ACG)} to address a class of AI-for-Science problems where the underlying causal structure is relatively simple, often taking the form of a tree. Our goal is to maintain a high likelihood for within-pair relationships throughout the entire diffusion process. To explain our algorithm more clearly, we use the paired relations \((A,B)\) and \((B,C)\) as a concrete example. The generalization to arbitrary contexts is detailed in Appendix~\ref{app:alg_details}. Notably, since the \((A,B)\) and \((B,C)\) pairs are trained jointly, they share some high-dimensional information associated with the common variable \(B\). Our objective is therefore twofold: not only to maximize the within-pair likelihood for each pair, but also to obtain a consistent estimate of \(B\) across pairs. Accordingly, in the following description, we introduce the method from the perspective of these shared high-dimensional features. The overall procedure consists of three stages: Consensus, Cooling, and Heating.

\subsection{Notations}
\label{sec:method_feature}
To rigorously characterize the interactions among these entities, we provide a semantic interpretation of how $A$, $B$, and $C$ evolve throughout the diffusion process; as discussed above, because training is performed via joint modeling, $A$ and $B$ share a set of common features within the $(A,B)$ pair, and $B$ and $C$ likewise share common features within the $(B,C)$ pair. We therefore decompose these vectors into shared and intrinsic features, $A=\{a,u\}$, $C=\{c,v\}$, and $B=\{b,u,v\}$, with the aim of preserving and maximizing the within-pair shared features $u$ and $v$—thereby maximizing the within-pair likelihood—while simultaneously obtaining a consensus estimate of the common variable $B$ across both pairs.

\subsection{Algorithm Workflow}
\paragraph{Training.}
For the diffusion model, we perform joint modeling of the paired data \((A,B)\) and \((B,C)\) by maximizing their corresponding likelihood objectives. After training, the diffusion model is able to generate \((A,B)\) pairs and \((B,C)\) pairs in an unconditional manner. And fortunately, foundation models for antigen–antibody generation (e.g., BoltzGen~\cite{boltzgen2025}) are already available; we can directly leverage these pretrained models without additional training to facilitate better design.

\paragraph{Inference.}
\textit{Initialization:} We run a set of diffusion processes in parallel. In practice, we use two copies of the model with identical weights. We independently sample two noise realizations and generate the pair \((A, B^{(A)})\) from the first copy and the pair \((B^{(C)}, C)\) from the second copy. Since these two branches start from different noise samples, they are initially incoherent, i.e., \(B^{(A)} \neq B^{(C)}\). Thus, when we initiate generation, we effectively obtain two parallel realizations of the same subject.

Our algorithm aims to eliminate the ambiguity arising from two parallel realizations of the same subject, $B^{(A)}$ and $B^{(C)}$, and to recover a single consistent $B$ as they co-evolve over diffusion time $t$. The detailed procedure is given below, which consists of three basic stages: Consensus, Heating, and Cooling. During this process, we maximize the common features \((u, v)\) defined above, with the dual goals of achieving consensus on \(B\) and maximizing the within-pair (conditional) log-likelihood.

\begin{algorithm}[ht]
\footnotesize
\caption{ACG: Annealed Co-Generation (Simplified $A, C$ Case)}
\label{alg:driver}
\KwIn{Contexts $A, C$, Schedules $\mathcal{S}_{heat}, \mathcal{S}_{sync}$}
\KwOut{Final Subject $B_{0}$}

$A_T, C_T, B_T^{(A)}, B_T^{(C)}, B_T^{(\emptyset)} \sim \mathcal{N}(0, \mathbf{I})$; \quad \textbf{Define} $\mathcal{X}_t = \{A_t, C_t, B_t^{(A)}, B_t^{(C)}, B_t^{(\emptyset)}\}$\;
\While{$t > 0$}{
    $t_{prev} \leftarrow t$; $(j_t, K_t, H) \leftarrow \mathcal{S}_{heat}(t)$;\quad $t_{target} \leftarrow \max(t - j_t, 0)$\;
    \For{$k \leftarrow 0$ \KwTo $K_t$}{
        \While{$t > t_{target}$}{
            \If{$t \in \mathcal{S}_{sync}$ \textbf{and} $k=0$}{
                \alglinehl{green!8}{$\mathcal{X}_{t}|_{B} \leftarrow \text{Consensus}(\mathcal{X}_t|_{B})$} \tcp*[r]{Phase 1: Force $B$-alignment (Break $u,v$)}
            }
            \alglinehl{blue!8}{$\mathcal{X}_{t-1} \leftarrow \text{Cooling}(\mathcal{X}_t, t)$} \tcp*[r]{Phase 3: Heal $u,v$}
            $t \leftarrow t - 1$\;
        }
        \If{$t > 0$ \textbf{and} $k < K$}{
            \alglinehl{red!8}{$\mathcal{X}_{t_{prev}} \leftarrow \text{Heating}(\mathcal{X}_t, t, t_{prev}, H)$} \tcp*[r]{Phase 2: Wash out corrupted averages}
            $t \leftarrow t_{prev}$\;
        }
    }
}
\Return $B_{0}$\;
\end{algorithm}

\paragraph{Phase 1: Consensus.}
In this stage, we force the parallel branches to agree on a single shared variable \(B\), typically via averaging:
\(B_{\mathrm{canon}} \leftarrow \text{Mean}(B^{(A)}, B^{(C)})\).
This operation yields a common estimate \(B_{\mathrm{canon}}=\{b_{\mathrm{new}},u_{\mathrm{new}},v_{\mathrm{new}}\}\), which connects the two pairs into a unified triplet \((A,B,C)\).
However, this common \(B_{\mathrm{canon}}\) may reduce the likelihood of both paired relations.

\paragraph{Phase 2: Heating.}
Because the newly generated pairs \((A,B)\) and \((B,C)\)—with variables \((a, b_{\text{new}}, u_{\text{new}}, v_{\text{new}})\) and \((b_{\text{new}}, u_{\text{new}}, v_{\text{new}}, c)\), respectively—may drift away from the previous high-likelihood region, we need a mechanism to recover from this deviation. The most naïve idea is simply to let the model try again; accordingly, we inject noise (i.e., add heat) via a backtracking step \(t \to t + j_t\), giving the system a fresh chance to search for a better configuration. Importantly, in this renewed attempt the optimization objective should place greater emphasis on maximizing the within-pair likelihood, which is precisely what the subsequent cooling phase is designed to enforce.

\paragraph{Phase 3: Cooling.}
Since the model is trained to maximize within-pair likelihood, it naturally tends to restore high-probability pairwise configurations. Concretely, if we take the post-consensus state for the \((A,B)\) branch, e.g., \((a, b_{\mathrm{new}}, u_{\mathrm{new}}, v_{\mathrm{new}})\), and continue running several denoising steps, the network will drive the variables toward regions that increase the likelihood of \((A,B)\). Figure~\ref{fig:vis} illustrates this behavior. As the system relaxes from \(t + j_t \to t\), the strong priors of the generative model re-impose the manifold constraints and regenerate interface features \((u, v)\) that are compatible with the contexts, improving within-pair likelihood.

By iterating this cycle, the system reaches a dynamic equilibrium in which \(B\) satisfies the geometric consensus \(B^{(A)} = B^{(C)}\), while simultaneously maximizing the within-pair likelihood for both \((A,B)\) and \((B,C)\). The detailed procedure is summarized in Algorithm~\ref{alg:driver}.

\subsection{Consensus Strategies}
\label{subsec:consensus_strat}

To achieve consensus in Phase 1, we derive two complementary strategies.
For simplicity of exposition, we focus here on the dual-context scenario ($A$ and $C$);
the generalization to arbitrary multi-target settings is detailed in Appendix~\ref{app:aggregation}.

\paragraph{Method 1: Joint Overlap Factorization.}
Assuming conditional independence, the joint distribution decomposes as:
\begin{equation}
    q(A,B,C) = \frac{q(A,B) q(B,C)}{q(B)}
\end{equation}
This yields a score-based consensus where each context contributes independently, with the unconditional prior subtracted to avoid double-counting\cite{zhang2023diffcollage}.

\paragraph{Method 2: Center-Weighted Fusion.}
The center variable $B$ is modeled as a weighted product of conditional influences:
\begin{equation}
    p(B|A,C) \propto p(B|A)^\beta p(B|C)^\alpha
\end{equation}
The resulting score is a weighted sum of conditional scores\cite{bar-tal2023multidiffusion}, with $\alpha = \beta = \frac{1}{2}$ as a common balanced choice.
Detailed derivations for both formulations are provided in Appendix~\ref{app:consensus_derivation}.
\subsection{Diffusion Scheduling}
When introducing the three stages---Consensus, Cooling, and Heating---it is also crucial to decide \emph{when} to invoke each stage and \emph{how} to alternate them over time. We therefore introduce two temporal mechanisms to schedule the process: one determines when to enforce consensus, and the other determines when to inject heat. At other timesteps, we deliberately avoid enforcing agreement and instead let the diffusion process generate each corresponding pair independently, relying on the model’s denoising dynamics to improve the within-pair likelihood.
\paragraph{Synchronization Schedule $\mathcal{S}_{sync}(t)$.}
We implement synchronization using an indicator function $\mathbb{I}_{sync}(t) \in \{0,1\}$. When $\mathbb{I}_{sync}(t)=1$, we apply the Consensus operator to align the intrinsic subject features $b$. Crucially, ACG follows a \textbf{``first-visit only''} policy: synchronization is enabled \textit{only} on the first cooling pass through timestep $t$. If the process reheats and later revisits the same $t$, we set $\mathbb{I}_{sync}(t)=0$.

\paragraph{Heating Schedule $\mathcal{S}_{heat}(t)$.}
To complement synchronization and mitigate potential structural artifacts, we use a heating schedule. Specifically, we define a mapping $t \mapsto (j_t, K_t, H)$ that controls the energy injection magnitude (jump size $j_t$), the exploration depth (number of resampling iterations $K_t$), and the \textit{heat height} $H \in (0,1]$, which scales the injected noise relative to the nominal jump:
\begin{equation}
    (j_t, K_t, H) =
    \begin{cases}
        (J_{heat}, K_{val}, H) & \text{if } t \in [T_{start}, T_{end}] \\
        (0, 0, 1) & \text{otherwise.}
    \end{cases}
\end{equation}
This formulation subsumes standard diffusion (where $j_t=0$) and resampling-based methods such as RePaint~\cite{lugmayr2022repaint}, which enables conditioned generation by iteratively reheating and resampling the denoising trajectory to satisfy local constraints using pretrained priors. By tuning $\mathcal{S}_{heat}$, we balance (i) removing corrupted interface averages introduced by $\mathcal{S}_{sync}$ and (ii) the additional computational cost from reheating and resampling.

By combining these two schedules, we can swap scheduling functions while keeping the core solver fixed, enabling fair and controlled comparisons. We also systematically analyze how different scheduling choices affect performance in our experiments. The specific configurations for Greedy, Consistent, and our proposed ACG are detailed in Section~\ref{sec:experiments}.
\section{Experiments}
\label{sec:experiments}
\begin{figure*}[t]
    \centering
    \includegraphics[width=0.9\linewidth]{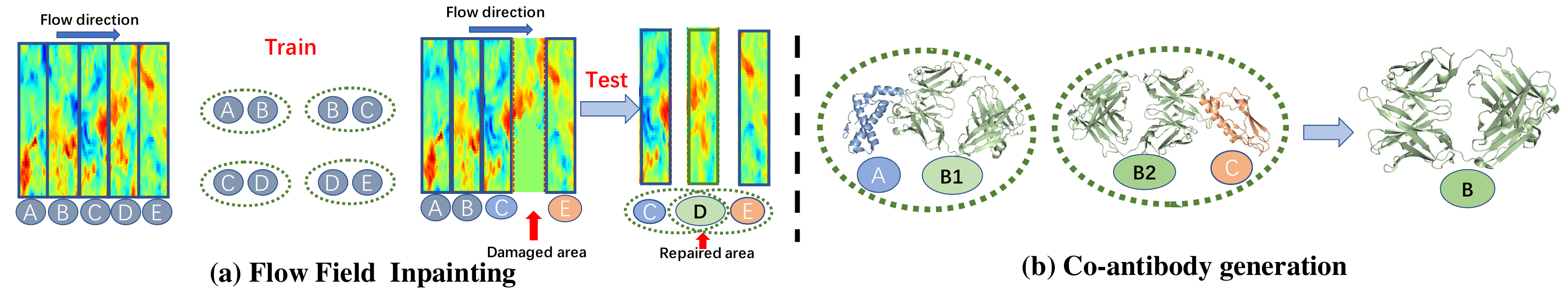}
    \caption{\textbf{Initialization for Different Tasks.} (a) Flow Field Inpainting: Based on the left-to-right flow, only adjacent regions are used for inpainting to maintain left–right consistency.
(b) Co-Antibody Generation: Using a pretrained model that jointly generates antigens and antibodies, we can design an antibody satisfying two antigens without additional training.}
    \label{fig:exp}
\end{figure*}

\paragraph{Overview.}
We validate our approach with two experiments (Fig.~\ref{fig:exp}): flow-field inpainting and multispecific antibody design.
For flow-field inpainting, we partition a large field into patches and strictly employ pairwise modeling. Since all variables share a coherent spatial grid, annealing is not required, demonstrating the efficacy of our pairwise formulation over conventional global modeling.
In contrast, for multispecific antibody design---where a single antibody must bind to multiple antigens---stronger constraints are needed to preserve conflicting pairwise interfaces. Here, we introduce annealing and evaluate our proposed \textbf{Annealed Co-Generation (ACG)} through ablations against standard pairwise baselines.

\vspace{-1em}
\paragraph{Method Comparison.} We instantiate the general schedules defined in Sec \ref{sec:method} to compare the following strategies. Beyond the single-target baseline, consensus methods fall into two paradigms based on \textit{when} coordination occurs: \textbf{Post-Hoc} methods first complete full generation trajectories then iteratively refine, while \textbf{In-Progress} methods interleave consensus within a single generation pass. For further details, please refer to the description in Fig.~\ref{fig:process_compare}.

\textit{Independent Oracle (No Consensus).} This represents the single-target lower bound where we generate subject $B$ conditioned solely on one context without any consensus operation.

\subparagraph{Post-Hoc Schedule.} To explore the system's behavior under macroscopic deviations, we employ a \textbf{global sawtooth schedule} that repeatedly cools to $t=0$ and reheats to progressively lower temperatures (e.g., $t=100, 50, 25$). Within this iterative regime, we evaluate three synchronization policies: \underline{No Synchronization} (branches evolve independently without interaction), \underline{Full Synchronization} (consensus is enforced at every timestep of every cycle), and \underline{Windowed Synchronization} (consensus is selectively applied only during the final cooling phase of each cycle).

\subparagraph{In-Progress Schedule.} These methods apply consensus during a single forward generation pass, optionally with local heating cycles for manifold recovery.

\textit{Greedy Consensus ($\mathcal{S}_{heat} \to \mathbf{0}, \mathbb{I}_{sync} \to 1$).} A standard diffusion trajectory adhering to consensus at every step ($\mathbb{I}_{sync}=1$) but without backtracking ($\mathcal{S}_{heat}=(0,0,1)$). This serves as the baseline for non-annealed joint generation.

\textit{Consistent  Consensus ($\mathcal{S}_{heat} \to \text{Active}, \mathbb{I}_{sync} \to 1$).} An intensive annealing regime where the process is repeatedly heated ($\mathcal{S}_{heat}=(J_{heat},K,H)$). Crucially, consensus is enforced at \textit{every} step ($\mathbb{I}_{sync}=1$) of both cooling and reheating, representing a "maximum constraint" regime that often over-constrains the interface features.

\textit{ACG (Ours) ($\mathcal{S}_{heat} \to \text{Active}, \mathbb{I}_{sync} \to \text{First-Visit}$).} Our proposed balanced schedule employs a "breathing" strategy ($\mathcal{S}_{heat}=(J_{heat}, K, H)$), where $H$ denotes the heat height scaling the noise injection magnitude. Unlike Consistent , ACG enforces consensus \textit{only during the initial cooling pass} ($\mathbb{I}_{sync}=\text{First-Visit}$), allowing branches to evolve freely during reheating to heal the shared feature subspaces $u$ and $v$. We distinguish consensus operators: variants denoted with \underline{/w} (e.g., ACG/w) utilize \underline{Center-Weighted Fusion}, while standard variants use \underline{Joint Overlap Factorization}.

\subsection{Flow field reconstruction}
\label{subsec:flow_recon}
\textbf{Dataset:} We collected velocity fields from $9$ flow cases and split the data into training and test sets at a $1\!:\!1$ ratio by frames, resulting in $7740$ frames for training and $7740$ frames for testing. Each velocity field has shape $H \times W \times 2 = 64 \times 160 \times 2$, where the two channels correspond to the streamwise and transverse velocity components $(u, v)$. More details about the datasets can be found in Appendix~\ref{app:dataset_field}. To construct supervised learning pairs, we randomly corrupt spatial regions within each frame using manually designed masking patterns, and use the original uncorrupted fields as labels. Since it is difficult to enumerate all possible corruption patterns in real-world scenarios, we apply different corruption schemes to the training and test sets to better emulate practical conditions and to evaluate the model's robustness and generalization.

For the pairwise patch modeling baseline, we partition each frame into patches of size $64 \times 32$, yielding $30960$ paired samples. Because the model takes a patch pair as a joint input, the effective input size is $64 \times 64$. Notably, the test set still contains a subset of uncorrupted patches, which are difficult to fully exploit when modeling the complete $64 \times 160$ field. In contrast, the pairwise patch formulation can leverage these uncorrupted patches, increasing the number of usable test pairs from $30960$ to $49511$.

\textbf{Zero-Shot Test:} To further assess the practical performance of different methods, we additionally collected $2$ entirely new cases (a total of $6464$ frames) as an extra test set. We directly apply the models pretrained in the first set of experiments to this dataset for inpainting, without any additional fine-tuning.

\label{subsec:flow_results}

\subsubsection{Results Analysis}
For all flow-field inpainting experiments, we report reconstruction quality using mean squared error (MSE), peak signal-to-noise ratio (PSNR), and the structural similarity index measure (SSIM), with all metrics computed directly in the original (physical) flow-field space without any additional normalization. For a fair comparison, all methods use the same UNet architecture~\cite{ronneberger2015u}, as shown in Table~\ref{tab:flow_field}.
Best results are highlighted in bold. Conventional inpainting baselines (e.g., UNet~\cite{ronneberger2015u}) struggle when the train--test distribution shift is large. RePaint~\cite{lugmayr2022repaint} improves adaptability by conditioning inference on the remaining (unmasked) region; however, due to the near single-manifold nature of flow fields, it redundantly models information and enlarges the search space.

In contrast, our ACG framework with Pairwise modeling with optimization models relationships between patch pairs, reducing the modeling space. Splitting one image into \(5\) patches yields four pairwise groups, effectively increasing the training data by \(4\times\) and substantially improving performance. In our basic setting, ACG refers to using DiffCollage~\cite{zhang2023diffcollage}, while ACG/w refers to the Center-Weighted Fusion scheme.Compared with the DiffCollage~\cite{zhang2023diffcollage} paradigm (Joint Overlap Factorization), Center-Weighted Fusion avoids two-stage training, mitigating forgetting when data are limited. Moreover, patch-level modeling can exploit the uncorrupted regions at test time, further boosting performance. Table~\ref{tab:flow_zero_shot} reports results on a completely new case, showing similar results, where our pairwise modeling achieves a clear improvement.

\begin{table}[t]
  \centering
  \caption{Experimental results for flow-velocity field inpainting.}
  \label{tab:flow_field}

  \setlength{\tabcolsep}{3.5pt}
  \renewcommand{\arraystretch}{1.05}
  \footnotesize

  \resizebox{0.95\linewidth}{!}{
  \begin{tabular}{l l c c c}
    \toprule
    \textbf{Model} & \textbf{Setting} &
    \textbf{MSE ($\downarrow$)} & \textbf{PSNR ($\uparrow$)} & \textbf{SSIM ($\uparrow$)} \\
    \midrule
    UNet~\cite{ronneberger2015u} & Case & 7.3232 & 16.270 & 0.3884 \\
    \midrule
    \multicolumn{5}{l}{\small\textbf{\textit{Joint modeling with optimization}}} \\
    RePaint~\cite{lugmayr2022repaint} & Case & 0.1878 & 32.388 & 0.6599 \\
    \midrule
    \multicolumn{5}{l}{\small\textbf{\textit{Pairwise modeling with optimization}}} \\
    ACG~\cite{ zhang2023diffcollage} & Patch     & 0.1320 & 34.003 & 0.7555 \\
    ACG/w~\cite{bar-tal2023multidiffusion} & Patch     & 0.1300 & 34.277 & 0.7448 \\
    ACG~\cite{zhang2023diffcollage} & All Patch & 0.1247 & 34.240 & 0.7629 \\
    \rowcolor{red!8}ACG/w~\cite{bar-tal2023multidiffusion} & All Patch & \textbf{0.1027} & \textbf{35.291} & \textbf{0.7868} \\
    \bottomrule
  \end{tabular}
  }
\end{table}

\begin{table}[t]
  \centering
  \caption{Zero-shot flow-field quantitative evaluation.}
  \label{tab:flow_zero_shot}

  \setlength{\tabcolsep}{3.5pt}
  \renewcommand{\arraystretch}{1.05}
  \footnotesize

  \resizebox{0.95\linewidth}{!}{
  \begin{tabular}{l l c c c}
    \toprule
    \textbf{Model} & \textbf{Setting} &
    \textbf{MSE ($\downarrow$)} & \textbf{PSNR ($\uparrow$)} & \textbf{SSIM ($\uparrow$)} \\
    \midrule
    UNet~\cite{ronneberger2015u} & Case & 1.5265 & 12.023 & 0.2381 \\
    \midrule
    \multicolumn{5}{l}{\small\textbf{\textit{Joint modeling with optimization}}} \\
    RePaint~\cite{lugmayr2022repaint} & Case & 0.2770 & 29.648 & 0.5555 \\
    \midrule
    \multicolumn{5}{l}{\small\textbf{\textit{Pairwise modeling with optimization}}} \\
    ACG~\cite{ zhang2023diffcollage} & Patch     & 0.2536 & 30.122 & 0.5831 \\
    ACG/w~\cite{bar-tal2023multidiffusion}& Patch     & \textbf{0.2443} & 30.249 & 0.5903 \\
    ACG~\cite{zhang2023diffcollage} & All Patch & 0.2496 & 30.195 & 0.5878 \\
    \rowcolor{red!8}ACG/w~\cite{bar-tal2023multidiffusion} & All Patch & 0.2447 & \textbf{30.250} & \textbf{0.5929} \\
    \bottomrule
  \end{tabular}
  }
\end{table}

\subsection{Multispecific Antibody Design}
\label{subsec:antibody_results}
\begin{figure*}[t!]
  \centering
  \begin{minipage}[t]{0.60\textwidth}
    \vspace{0pt} 
    \centering
    \captionsetup{width=0.98\linewidth}
    \footnotesize
    \setlength{\tabcolsep}{3.5pt}
    \renewcommand{\arraystretch}{1.1} 
    \captionof{table}{Ablation results on Split-A. \textit{Top}: Comparison between unsynchronized methods and \textit{In-Progress} schedules. Middle: Performance of the ACG method under two different parameter settings, specifically $K$ and Heat Height. Bottom: Performance of the \textit{Post-Hoc} Schedule.}
    \label{tab:ablation_selected}

    \resizebox{\linewidth}{!}{
    \begin{tabular}{lcccccc}
      \toprule
      \multirow{2}{*}{\textbf{Configuration}} & \multicolumn{3}{c}{\textbf{RMSD} (\text{\AA}) ($\downarrow$)} & \multicolumn{3}{c}{\textbf{design-ipTM} ($\uparrow$)} \\
      \cmidrule(lr){2-4} \cmidrule(lr){5-7}
      & Ag. A & Ag. C & Avg. & Ag. A & Ag. C & Avg. \\
      \midrule
      \multicolumn{7}{l}{\textbf{\textit{Method Comparison}}} \\
      Single-target \cite{boltzgen2025} & 4.11 & 3.85 & 3.98 & 0.18 & \textbf{0.21} & 0.20 \\
      Greedy \cite{zhang2023diffcollage} & 4.45 & 4.48 & 4.47 & 0.17 & 0.19 & 0.18 \\
      Greedy/w \cite{bar-tal2023multidiffusion} & 3.98 & 4.06 & 4.02 & 0.18 & 0.19 & 0.19 \\
      Consistent \cite{liu2022compositional} & 4.64 & 4.69 & 4.67 & 0.18 & 0.17 & 0.18 \\
      Consistent /w \cite{bar-tal2023multidiffusion}& 4.46 & 4.41 & 4.44 & 0.18 & 0.19 & 0.19 \\
      ACG & 3.75 & 3.78 & 3.76 & 0.19 & 0.20 & \textbf{0.20} \\
      \rowcolor{red!8} ACG/w & \textbf{3.75} & \textbf{3.77} & \textbf{3.76} & \textbf{0.19} & 0.19 & 0.19 \\
      \midrule
      \multicolumn{7}{l}{\textbf{\textit{Exploration Depth ($K$)}} \cite{lugmayr2022repaint}} \\
      $K=1$ & 3.76 & 3.77 & 3.76 & 0.19 & 0.19 & 0.19 \\
      $K=2$  & \textbf{3.75} & \textbf{3.77} & \textbf{3.76} & \textbf{0.19} & \textbf{0.19} & \textbf{0.19} \\
      $K=3$ & 3.88 & 3.95 & 3.92 & 0.19 & 0.19 & 0.19 \\
      $K=4$ & 3.91 & 3.86 & 3.89 & 0.18 & 0.18 & 0.18 \\
      \midrule
      \multicolumn{7}{l}{\textbf{\textit{Noise Scale (Heat Height)}} \cite{song2019generative, song2020score}} \\
      Height 1.00 & \textbf{3.75} & 3.77 & \textbf{3.76} & \textbf{0.19} & \textbf{0.19} & \textbf{0.19} \\
      Height 0.75 & 3.88 & \textbf{3.72} & 3.80 & 0.19 & 0.19 & 0.19 \\
      Height 0.50 & 3.83 & 3.78 & 3.80 & 0.19 & 0.19 & 0.19 \\
      \midrule
      \multicolumn{7}{l}{\textbf{\textit{Post-Hoc Schedule}} \cite{yu2023freedom, bansal2024universal}} \\
      Full Step Sync. & 3.87 & 3.90 & 3.89 & \textbf{0.19} & 0.19 & 0.19 \\
      Windowed Sync. & \textbf{3.77} & \textbf{3.80} & \textbf{3.79} & 0.18 & 0.19 & \textbf{0.19} \\
      No Sync. & 4.14 & 4.01 & 4.07 & 0.18 & \textbf{0.20} & 0.19 \\
      \bottomrule
    \end{tabular}
    }
  \end{minipage}\hfill
  \begin{minipage}[t]{0.40\textwidth}
    \vspace{0pt} 
    \centering
    \captionsetup{width=0.98\linewidth}
    \vspace{2.8em} 
    \includegraphics[width=\linewidth]{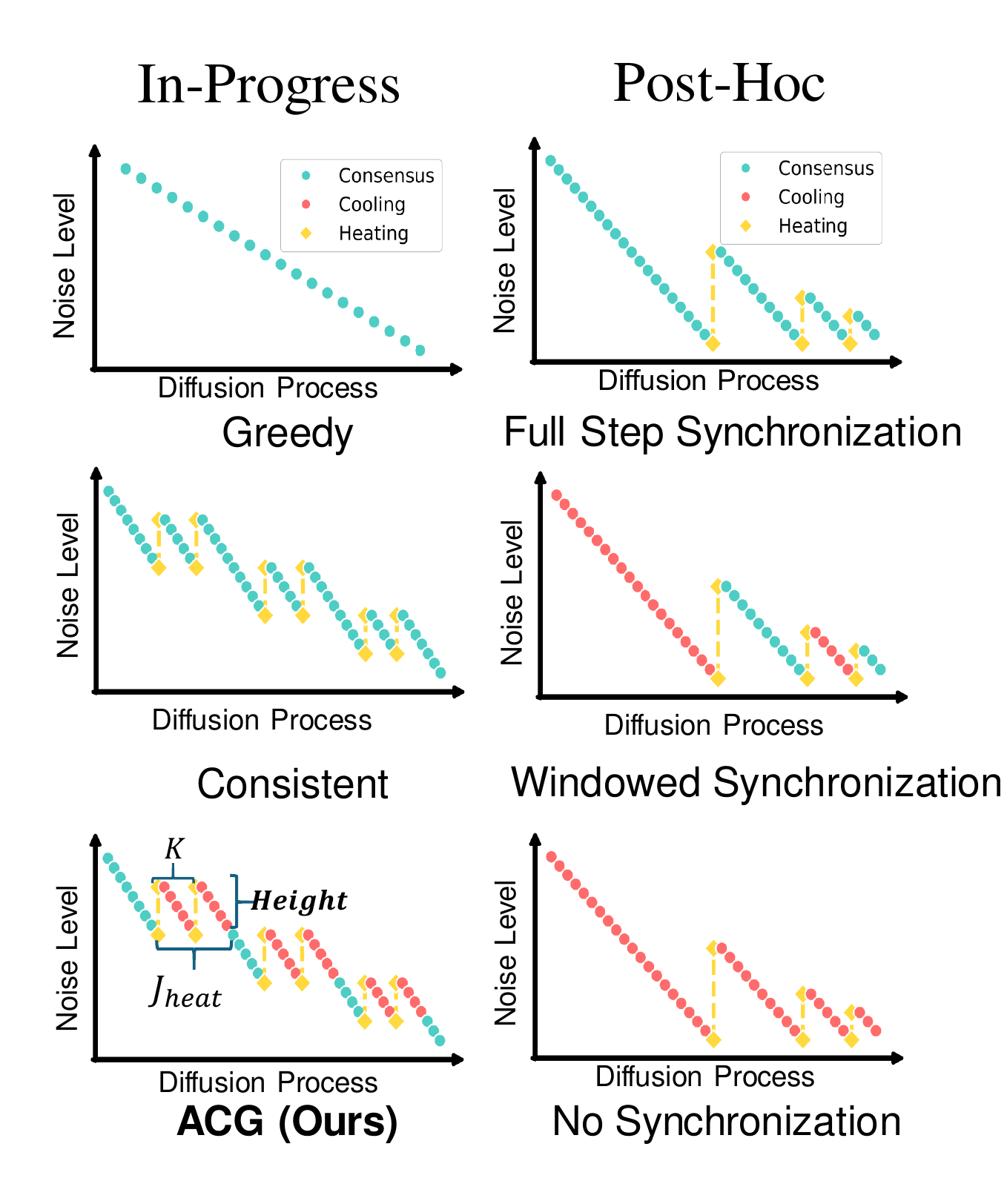}
    \vspace{-0.6em} 
    \captionof{figure}{Illustration of scheduling strategies for different settings. $K$ denotes the exploration depth (resampling iterations), Height represents the maximum noise level during reheating, and $J_{heat}$ indicates the step size for energy injection.}
    \label{fig:process_compare}
  \end{minipage}
  \vspace{-0.5em}
\end{figure*}
\textbf{Dataset:} To evaluate \textbf{multi-specific antibody design} (one antibody binding two distinct antigens), we curated a benchmark from SAbDab~\cite{sabdab}. Following rigorous filtering protocols from prior works~\cite{luo2022antigen, kong2022conditional}, we identified ``ground truth'' multi-specificity cases by grouping antigens bound by antibodies with identical heavy and light chain sequences. This yielded $80$ homologous antigen pairs (double-antigen clusters). More details about the datasets can be found in Appendix~\ref{app:dataset_details}. We constructed two evaluation subsets: \textbf{Split-A} consists of the $10$ smallest clusters, designed to minimize computational overhead for rapid hyperparameter tuning and ablation studies. \textbf{Split-B} comprises the $10$ clusters where the single-target oracle (BoltzGen) achieves its best performance. This selection establishes a high-quality baseline, ensuring that our evaluation focuses on the challenge of joint structural compatibility in realistic, high-fidelity scenarios rather than being confounded by basic generation failures.

\subsubsection{Comparative Analysis}

\label{subsubsec:antibody_ablations}

To evaluate the effectiveness of our approach, we provide a comprehensive comparison with several methods.
For a fair comparison, all methods utilize the standardized BoltzGen pipeline~\cite{boltzgen2025} for inverse folding, structural refolding, and interface analysis to assess candidate quality.
For unpaired antibodies (where the antibody is generated based solely on antigen A, Ag.A, while antigen C, Ag.C, serves as a reference), we calculated the metrics for the generated antibody against Ag.A and Ag.C separately, and averaged the two values.

\textbf{Metric Selection Rationale.} It is important to note that consensus-based optimization primarily acts on the internal geometry of the generated subject to resolve conflicting constraints, rather than altering the macroscopic docking pose relative to the antigens. Consequently, \textbf{RMSD} (\AA)---which measures structural deviation and stability---is a more sensitive and reliable indicator of method performance in this context compared to \textbf{design-ipTM}, which evaluates interface packing distance and often shows lower variance across methods~\cite{evans2021protein, yin2022benchmarking}. Thus, our analysis focuses primarily on RMSD improvements.

We first utilize the efficiency-driven Split-A subset to identify optimal thermodynamic configurations. Table~\ref{tab:ablation_selected} presents the comparative analysis.
First, we establish a baseline using the \textit{Single-target} approach (standard BoltzGen), which achieves an average RMSD of 3.98~\AA.
Naive attempts to introduce multi-target constraints fail to improve upon this baseline: the \textit{Greedy} variants (no heating) are trapped in local minima (RMSD 4.47--4.02~\AA), while the \textit{Consistent } strategy—which enforces consensus at every cooling step—over-constrains the branches, preventing relaxation onto the natural protein manifold and resulting in the worst performance (RMSD $>$4.4~\AA).

\textit{ACG} outperforms both naive strategies and the \textit{Single-target} baseline, achieving the highest precision (RMSD 3.76\,\AA). Its ``First-Visit'' policy enforces consensus only during the initial cooling pass, allowing free exploration during reheating to resolve structural artifacts. Notably, the specific consensus formulation is less critical than its synergy with annealing; both standard overlap and weighted fusion (\textit{/w}) yield comparable results. This robustness suggests that \textit{ACG} succeeds through its scheduling rather than the fusion metric, with $K=2$ iterations striking the optimal balance between exploration and refinement.

Finally, we analyze the Post-Hoc Schedule (Table~\ref{tab:ablation_selected}, bottom). We observe that \textit{No Synchronization} results in poor structural alignment (RMSD 4.07~\AA), confirming the necessity of coordination. However, \textit{Full Step Synchronization} fails to achieve optimal results (RMSD 3.89~\AA), likely by restricting the trajectory's ability to relax into valid protein conformations. In contrast, \textit{Windowed Synchronization} yields the best performance (RMSD 3.79~\AA), supporting our hypothesis that consensus should be applied strategically rather than continuously~\cite{yu2023freedom, bansal2024universal}.

\subsubsection{Multi-Target Binding Performance}
\label{subsubsec:antibody_main_results}

Table~\ref{tab:antibody_results} presents the aggregate performance on Split-B.
The \textit{Single-target} exhibits severe performance degradation when evaluated on the paired target, highlighting the inherent difficulty of the joint constraint.
Naive joint optimization strategies offer some improvement in compatibility but still fall short of the precision required for therapeutic candidates.
In distinct contrast, our ACG method achieves the lowest average RMSD.
This represents a significant improvement over the independent baseline, confirming that explicit coordination via consensus successfully resolves structural conflicts.
\begin{table}[t]
\centering
\scriptsize
\setlength{\tabcolsep}{2pt}

\renewcommand{\arraystretch}{0.95}
\caption{\textbf{Results on baselines under a more realistic design setting.}
Main results for multi-specific antibody co-design on the Split-B benchmark. }
\label{tab:antibody_results}
\resizebox{\columnwidth}{!}{
\begin{tabular}{lcccccc}
\toprule
\multirow{2}{*}{\textbf{Method}} & \multicolumn{3}{c}{\textbf{RMSD} (\AA) (\(\downarrow\))} & \multicolumn{3}{c}{\textbf{design-ipTM} (\(\uparrow\))} \\
\cmidrule(lr){2-4} \cmidrule(lr){5-7}
 & Ag. A & Ag. C & Avg. & Ag. A & Ag. C & Avg. \\
\midrule
Single-target \cite{boltzgen2025} & 3.21 & 3.38 & 3.30 & \textbf{0.37} & 0.29 & 0.33 \\
Greedy \cite{zhang2023diffcollage} & 3.87 & 3.86 & 3.87 & 0.34 & 0.33 & 0.34 \\
Greedy/w \cite{bar-tal2023multidiffusion} & 3.27 & 3.33 & 3.30 & 0.33 & 0.31 & 0.32 \\
Consistent  \cite{liu2022compositional} & 4.22 & 4.21 & 4.22 & 0.35 & 0.34 & 0.35 \\
Consistent /w \cite{bar-tal2023multidiffusion} & 3.15 & 3.13 & 3.14 & 0.32 & 0.32 & 0.32 \\
ACG~\cite{zhang2023diffcollage} & 3.75 & 3.70 & 3.73 & 0.35 & \textbf{0.34} & \textbf{0.35} \\
\rowcolor{red!8} ACG/w\cite{bar-tal2023multidiffusion} & \textbf{2.96} & \textbf{3.02} & \textbf{2.99} & 0.33 & 0.33 & 0.33 \\
\bottomrule
\end{tabular}
}
\end{table}
Notably, our joint optimization results even surpass the average performance of the single-target oracle itself. This suggests that the consensus mechanism acts as a regularizer, filtering out spurious conformational states that may exist in isolated single-target generation.
As detailed in Table~\ref{tab:antibody_results_per_target}, this advantage is particularly pronounced on structurally divergent pairs, where ACG achieves sub-3.0~\AA\ precision, demonstrating robust generalization capabilities.

\subsubsection{Analysis of the Cooling Process}
\begin{figure}[H]
    \centering
    \includegraphics[width=0.85\linewidth]{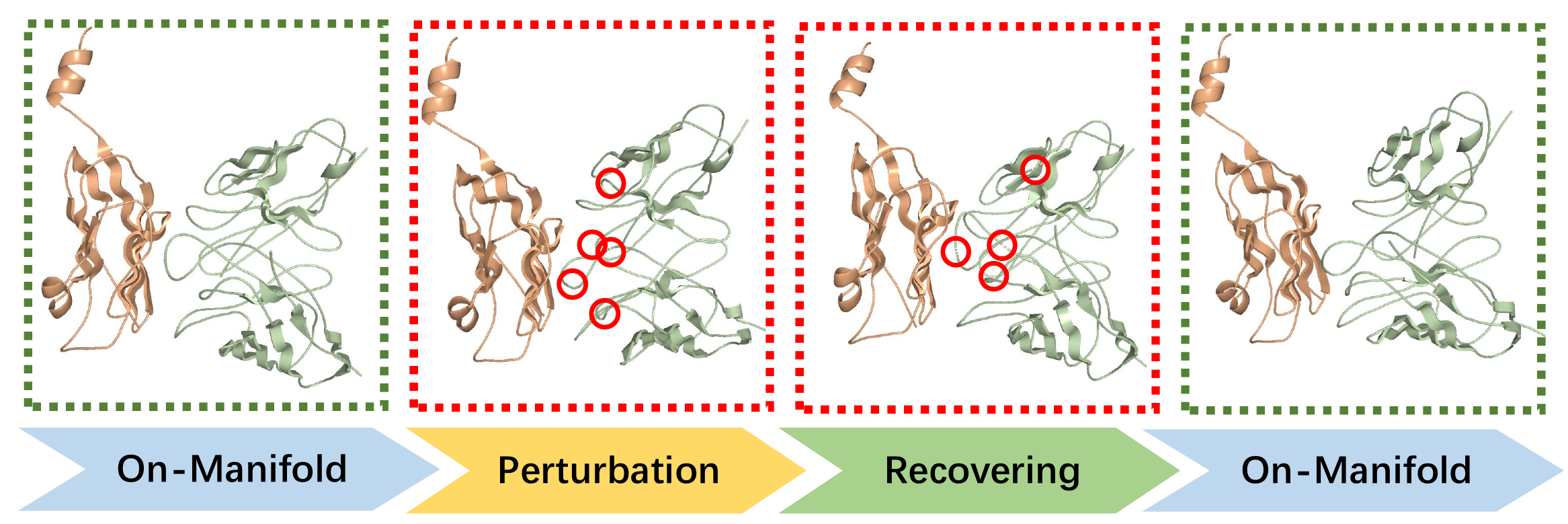}
    \small
    \caption{\textbf{Geometric Repair Analysis.} Joint optimization may temporarily distort local geometry. The diffusion process subsequently corrects these inconsistencies and resolves steric clashes, restoring structural validity on the pretrained manifold.}
    \label{fig:vis}
\end{figure}

\noindent\textbf{Cooling as Structural Repair.} The cooling stage in ACG mitigates structural violations induced by the consensus operator. Forcing a single antibody to satisfy conflicting antigen geometries compromises \textbf{intrinsic geometric constraints} (bond lengths/angles), causing \textbf{steric clashes} and unfavorable backbone conformations. This pushes the trajectory off the ``within-pair manifold'' into high-energy states. Cooling acts as a relaxation mechanism, restoring these intermediates to a low-energy protein manifold.

As shown in Fig.~\ref{fig:vis}, BoltzGen~\cite{boltzgen2025} illustrates this recovery. Early consensus optimization induces geometric distortions, which the denoising dynamics (as $t \to 0$) subsequently relax. This cooling phase corrects internal errors, repacking side-chains and optimizing interfacial contacts to resolve clashes. By interleaving consensus updates with cooling, ACG guides the trajectory back to a low-energy, physically valid manifold, ensuring multi-specific antibodies are structurally sound and stable.
\section{Conclusion}
In this paper, we advocate pairwise block modeling rather than jointly modeling all variables. For the flow-field inpainting task, this formulation not only reduces the effective modeling space, but also leverages the complete regions as much as possible, thereby increasing the amount of usable training signal. For co-antibody generation, our Annealed Co-Generation (ACG) framework enables, at inference time and without retraining the pretrained BoltzGen~\cite{boltzgen2025} model, the direct generation of antibodies that satisfy constraints imposed by both antigens while preserving the key structural characteristics of each antigen–antibody pair.

\bibliography{bib/string.bib,bib/vision.bib,bib/graphics.bib,bib/optim.bib,bib/learning.bib,bib/new.bib}
\bibliographystyle{unsrt}

\onecolumn
\appendix
\section{Dataset Construction Details}

\subsection{Flow Field Dataset}
\label{app:dataset_field}
For the generation of the flow field dataset, we utilized the GPU-accelerated large eddy simulation (LES) code, WiRE-LES, developed at EPFL's Wind Engineering and Renewable Energy (WiRE) Laboratory. This well-validated code has been extensively applied in numerous studies of atmospheric boundary layer (ABL) flows, particularly those involving wind turbines~\cite{porte2011wind, wu2012atmospheric, wu2015modeling, shamsoddin2017large, revaz2021large, vahidi2022physics, lin2024effects}. Within this framework, key inflow parameters---such as friction velocity and aerodynamic surface roughness length---can be prescribed to produce inflow conditions representative of diverse terrain types, ranging from flat homogeneous terrain to forest canopies.

\subsection{Co-antibody Datase}

\label{app:dataset_details}

In this section, we provide the specific details of our data curation pipeline, including source filtering, processing, and the exact algorithms used for clustering.

\subsubsection{The Challenge of Data Scarcity}
We analyzed the Structural Antibody Database (SAbDab) \cite{sabdab} following standard protocols \cite{luo2022antigen, kong2022conditional}. Our analysis confirms a severe scarcity of multi-specific signals: 88.4\% of epitope clusters are singletons, and 81.5\% of antibodies bind only a single unique antigen. This necessitates a generative approach capable of zero-shot inference \cite{chai2024zeroshot, watson2023rfdiffusion}, as sought after in recent models like Boltz-1 \cite{wong2024boltz1}, BoltzGen \cite{boltzgen2025}, and others \cite{yang2025repurposing}.

\subsubsection{Fragility of Protein Validation}
Current computational validation of de novo designs typically relies on a complex multi-stage pipeline: (1) structural generation via design models such as RFDiffusion \cite{watson2023rfdiffusion}, FrameDiff \cite{yim2023framediff}, or Chroma \cite{ingraham2023chroma}; (2) sequence design via inverse folding models like ProteinMPNN \cite{dauparas2022robust} or PiFold \cite{gao2022pifold}; and (3) structural verification via folding models such as AlphaFold2/3 \cite{jumper2021highly, abramson2024alphafold3}, Boltz-1 \cite{wong2024boltz1}, or specialized models like ProteinX and Boltz-2. This lengthy chain introduces cumulative errors and significant computational overhead, making the assessment of cross-reactive antibodies particularly challenging. Since our objective is to characterize the joint conditional distribution $P(y | x_1, x_2)$ for antibodies $y$ binding antigens $x_1$ and $x_2$ simultaneously, we prioritize structural consistency and joint binding potential. Our method aims to match the performance of direct single-target designs while significantly outperforming antigen-exchanged baselines, where naive single-target models ($y_1 \sim P(y|x_1)$ and $y_2 \sim P(y|x_2)$) fail to generalize to the cross-pairings $(x_1, y_2)$ and $(x_2, y_1)$.

\subsubsection{Data Source and Pre-processing}
We utilized the Structural Antibody Database (SAbDab) as our primary source of antibody-antigen complexes. To ensure high data quality suitable for geometric deep learning, we applied the following rigorous filtering criteria:

\begin{itemize}
    \item \textbf{Protein-Only Interactions:} We filtered for entries where the antigen type is strictly protein, excluding nucleic acids or small molecule ligands.
    \item \textbf{Resolution Cutoff:} Structures with a resolution worse than $4.5$\AA\ were discarded to remove low-quality experimental data. Structural comparisons use RMSD computed via Kabsch superposition \cite{kabsch1976solution}.
    \item \textbf{Completeness:} We required all CDR loops (CDR-H1, H2, H3, L1, L2, L3) to be fully resolved without missing residues.
    \item \textbf{Chain Integrity:} Heavy chains were limited to $\le 113$ residues and light chains to $\le 109$ residues following Chothia numbering constraints.
    \item \textbf{Size Filtering:} Antigens with a total sequence length exceeding 1,500 amino acids were excluded to remove excessively large protein complexes.
\end{itemize}

This initial filtering yielded the core dataset of valid antibody-antigen interfaces used for subsequent analysis.

\subsubsection{Clustering Strategy}
We employed a clustering strategy to construct our benchmark.

This strategy focuses on the functional aspect of multi-specificity, grouping antigens that are bound by the exact same antibody.
\begin{itemize}
    \item \textbf{Definition:} Two targets are clustered if they are bound by antibodies with identical heavy (VH) and light (VL) chain sequences.
    \item \textbf{Metric:} Exact string matching of the amino acid sequences of the VH and VL domains.
    \item \textbf{Rationale:} This captures "ground truth" multi-specificity, where a single biological antibody sequence has been experimentally observed to bind multiple distinct antigen structures.
\end{itemize}

\subsubsection{Efficiency-Driven Subset for Ablation (Split-A)}
To facilitate extensive hyperparameter optimization and ablation studies, we identified a high-efficiency subset of the double-antigen clusters, denoted as \textbf{Split-A}. This subset consists of the \textbf{10 smallest} antigen clusters selected based on total amino acid sequence length. This selection minimizes the computational resource requirements (GPU memory and inference time) for each sampling iteration, enabling the evaluation of over 30 distinct algorithmic configurations (see Appendix~\ref{app:ablation_results}). Despite its reduced size, Split-A maintains structural diversity and serves as a reliable proxy for the broader dataset in characterizing the impact of consensus hyperparameters. The target antigen pairs in Split-A, represented by their PDB identifiers, are: \texttt{5dhy/5dhv}, \texttt{6j5d/6j5f}, \texttt{2dqj/1c08}, \texttt{6ba5/6ban}, \texttt{6mhg/6mb3}, \texttt{8yat/8k33}, \texttt{8pnv/8pnu}, \texttt{9mi0/9mib}, \texttt{8hbj/8hbg}, and \texttt{9cfh/9cfg}.

\subsubsection{Benchmark Selection via BoltzGen (Split-B)}
To establish a robust evaluation set for multi-target generation, we further filtered the 80 identified double-antigen clusters. We selected the \textbf{10 clusters} where the single-target oracle (BoltzGen \cite{boltzgen2025}) achieves the best performance in terms of average RMSD, denoted as \textbf{Split-B}. By prioritizing cases with the most reliable single-target reference designs, we ensure a high-quality "gold standard" to evaluate the framework's ability to maintain or enhance binding precision during joint optimization. This selection process yields a total of \textbf{10} double-antigen clusters for our main evaluation. The target antigen pairs, denoted by their PDB identifiers, are: \texttt{4jre/4jr9}, \texttt{7zf9/7bei}, \texttt{5tru/7elx}, \texttt{4yxk/4h88}, \texttt{2r4r/2r4s}, \texttt{9ptm/9me5}, \texttt{7lkh/7lkf}, \texttt{7x1m/8hwt}, \texttt{6j5d/6j5f}, and \texttt{5tr1/5tqq}.

\subsubsection{Experimental Settings and Hyperparameters}
We strictly followed the pipeline provided by BoltzGen (\url{https://github.com/Zhan-Tao/BoltzGen}) for both the design and validation phases. Specifically, we used the default hyperparameters for diffusion sampling, structure prediction (using Boltz-1), and interface evaluation. Our multi-target framework was implemented as a plug-and-play modification to the BoltzGen sampling loop, maintaining the underlying model weights and basic structural constraints unchanged.
\section{Additional Algorithmic Details}
\label{app:alg_details}

In this section, we provide the detailed sub-routines and reference algorithms used in our framework, generalized to the case of $N$ arbitrary target contexts $\mathcal{C} = \{C_k\}_{k=1}^N$.

Algorithm~\ref{alg:driver_generalized} outlines the top-level orchestration for multi-target generation, extending the simplified case presented in the main text to arbitrary $N$ branches plus one unconditional branch.

\begin{algorithm}[ht]
\caption{ACG: Generalized Annealed Co-Generation ($N$-Target)}
\label{alg:driver_generalized}
\KwIn{Contexts $\mathcal{C} = \{C_k\}_{k=1}^N$, Mask $\Mask$, Schedules $\mathcal{S}_{heat}, \mathcal{S}_{sync}$}
\KwOut{Final States $\mathcal{X}_{0} = \{\State_0^{(1)}, \dots, \State_0^{(N+1)}\}$}

$\mathcal{X}_T \sim \mathcal{N}(0, \mathbf{I})$; $k \leftarrow 0$; \quad \textbf{Init:} Parallel branches sharing subject region\;
\While{$t > 0$}{
    $t_{prev} \leftarrow t$; $(j_t, K_t, H) \leftarrow \mathcal{S}_{heat}(t)$;\quad $t_{target} \leftarrow \max(t - j_t, 0)$\;
    \For{$k \leftarrow 0$ \KwTo $K_t$}{
        \While{$t > t_{target}$}{
            \If{$t \in \mathcal{S}_{sync}$ \textbf{and} $k=0$}{
                \alglinehl{green!8}{$\mathcal{X}_{t} \leftarrow \text{Consensus}(\mathcal{X}_t, t, \mathcal{C}, \Mask)$}
            }
            \alglinehl{blue!8}{$\mathcal{X}_{t-1} \leftarrow \text{Cooling}(\mathcal{X}_t, t, \mathcal{C})$};\\
            $t \leftarrow t - 1$\;
        }
        \If{$t > 0$ \textbf{and} $k < K_t$}{
            \alglinehl{red!8}{$\mathcal{X}_{t_{prev}} \leftarrow \text{Heating}(\mathcal{X}_t, t, t_{prev}, H)$};\\
            $t \leftarrow t_{prev}$\;
        }
    }
}
\Return $\mathcal{X}_{0}$\;
\end{algorithm}

\paragraph{Synchronized Sampling Step}
Algorithm~\ref{alg:step_update} defines a single reverse diffusion transition \cite{Ho20a, Song20c} that aligns multiple branches. Given current states $\State_t^{(1..N+1)}$, timestep $t$, contexts $\{C_k\}$, and mask $\Mask$, it first predicts clean states $\hat{\State}^{(k)}_0$ in parallel. A canonical consensus $S_{canon}$ is then calculated to synchronize the subject part across branches, which is subsequently injected back into each individual branch's prediction while retaining local context. This ensures that the collective subject evolves consistently while satisfying individual constraints.

\begin{algorithm}[ht]
\SetAlgoLined
\DontPrintSemicolon
\caption{SynchronizedStep}
\label{alg:step_update}

\KwIn{Current State $\mathcal{X}_t$, Time $t$, Contexts $\mathcal{C}$, Mask $\Mask$}
\KwOut{Perturbed State $\mathcal{X}_{t}'$}

\tcp{1. Extract Subject States}
$B_t^{(k)} \leftarrow \mathcal{X}_t^{(k)}[\Mask]$ for $k = 1, \ldots, N+1$\;

\tcp{2. Compute Canonical Consensus (Alg. \ref{alg:aggregation})}
$B_{canon} \leftarrow \text{UnifiedAggregation}(\{B_t^{(k)}\}, \Mask, \lambda)$\;

\tcp{3. Consensus Injection (Perturbation)}
\For{$k \leftarrow 1$ \KwTo $N+1$}{
    \tcp{Force Consensus on Subject Region}
    $\mathcal{X}_t^{(k)}[\Mask] \leftarrow B_{canon}$\;
    \tcp{Context/Interface Regions remain untouched}
}

\Return $\mathcal{X}_{t}$\;
\end{algorithm}

\subsection{Theoretical Derivation: Consensus Strategies}
\label{app:consensus_derivation}

In this section, we derive two complementary consensus strategies referenced in Section~\ref{subsec:consensus_strat}.

\subsubsection{Method 1: Joint Overlap Factorization (DiffCollage-style)}

Let $U=(A,B,C)$, where $(A,B)$ is observed (or generated first) and $C$ is completed given $B$.

Assume conditional independence:
\begin{equation}
q(C\mid A,B)=q(C\mid B).
\end{equation}
Then
\begin{equation}
q(A,B,C) = \frac{q(A,B)\,q(B,C)}{q(B)}.
\end{equation}

The score (gradient of the log-density) decomposes as
\begin{align}
\nabla_U \log q(U)
&= \nabla_U\Bigl(\log q(A,B)+\log q(B,C)-\log q(B)\Bigr) \nonumber \\
&= \nabla_U \log q(A,B)+\nabla_U \log q(B,C)-\nabla_U \log q(B).
\end{align}

Equivalently, written component-wise:
\begin{align}
\nabla_A \log q(U) &= \nabla_A \log q(A,B), \\
\nabla_B \log q(U) &= \nabla_B \log q(A,B)+\nabla_B \log q(B,C)-\nabla_B \log q(B), \\
\nabla_C \log q(U) &= \nabla_C \log q(B,C).
\end{align}

This justifies subtracting the unconditional score $\nabla_B \log q(B)$ to avoid double-counting the prior when fusing pairwise models.

\subsubsection{Method 2: Center-Weighted Fusion}

To model how the neighbors $(A,C)$ influence the center variable $B$, we use a weighted product:
\begin{equation}
p(B\mid A,C)\propto p(B\mid A)^{\beta}\,p(B\mid C)^{\alpha}.
\end{equation}

Equivalently, introducing a normalizer $Z(A,C)$,
\begin{equation}
\log p(B\mid A,C)=\beta\log p(B\mid A)+\alpha\log p(B\mid C)-\log Z(A,C).
\end{equation}

The conditional score (with respect to $B$) is a weighted sum:
\begin{equation}
\nabla_B \log p(B\mid A,C)
= \beta\,\nabla_B \log p(B\mid A)+\alpha\,\nabla_B \log p(B\mid C),
\end{equation}
since $Z(A,C)$ does not depend on $B$. A common special case is $\alpha=\beta=\tfrac{1}{2}$, which corresponds to averaging the conditional scores.

\subsection{Unified Multi-Target Aggregation}
\label{app:aggregation}

Algorithm~\ref{alg:aggregation} implements a practical fusion formula that generalizes the above strategies. Given $N$ conditional predictions $\{B^{(k)}\}_{k=1}^N$ and one unconditional baseline $B^{(\emptyset)}$, it computes the canonical consensus:
\begin{equation}
\label{eq:aggregation_appendix}
B_{canon} = \mu_{cond} + \lambda \cdot (\mu_{cond} - B^{(\emptyset)})
\end{equation}
where $\mu_{cond} = \frac{1}{N}\sum_{k=1}^{N} B^{(k)}$ is the mean conditional prediction and $\lambda$ is the guidance scale.

\begin{algorithm}[ht]
\SetAlgoLined
\DontPrintSemicolon
\caption{UnifiedAggregation (Mean \& Product-of-Experts)}
\label{alg:aggregation}

\KwIn{Subjects $\{B^{(k)}\}_{k=1}^{N+1}$, Mask $\Mask$, Guidance $\lambda$}
\KwOut{Canonical Subject $B_{canon}$}

\tcp{Guided Fusion}
$\mu_{cond} \leftarrow \text{Mean}(B^{(1 \dots N)})$ \tcp*{Average of Conditional Branches}
$u_{base} \leftarrow B^{(N+1)}$ \tcp*{Unconditional Baseline}

\tcp{$\lambda = 0$: Mean (Method 2, $\alpha=\beta=\frac{1}{2}$); $\lambda = N-1$: Joint Overlap (Method 1)}
$B_{canon} \leftarrow \mu_{cond} + \lambda \cdot (\mu_{cond} - u_{base})$\;

\Return $B_{canon}$\;
\end{algorithm}

\paragraph{Connection to Theoretical Methods.}
When $\lambda = 0$, we recover the Mean Aggregation strategy, which corresponds to Method 2 with $\alpha = \beta = \frac{1}{2}$ (balanced center-weighted fusion). When $\lambda = N-1$, we recover the full score subtraction of Method 1 (joint overlap factorization). In practice, $\lambda$ serves as a hyperparameter to balance diversity (low $\lambda$) against constraint satisfaction (high $\lambda$).

\section{Notation and Definitions}
\label{app:notation}
\begin{table}[H]
    \centering
    \caption{Summary of Notation and Symbols used in our framework.}
    \label{tab:notation}

    \renewcommand{\arraystretch}{1.2}
    \begin{tabularx}{\linewidth}{l X}
    \toprule
    \textbf{Symbol} & \textbf{Description} \\
    \midrule
    \multicolumn{2}{l}{\textit{Core Variables (Main Text)}} \\
    $B$ & \textbf{Subject:} The entity being designed (e.g., antibody sequence or structure). \\
    $A, C$ & Example contexts (targets) used in the main text. \\
    $B^{(A)}, B^{(C)}$ & Branch-specific predictions of the subject. \\
    \midrule
    \multicolumn{2}{l}{\textit{Feature Decomposition}} \\
    $b$ & Intrinsic features of the subject $B$ (its identity or style). \\
    $a, c$ & Features intrinsic to the contexts $A$ and $C$ (irrelevant to $B$). \\
    $u$ & Feature subspace shared by $A$ and $B$ (interface features). \\
    $v$ & Feature subspace shared by $C$ and $B$ (interface features). \\
    \midrule
    \multicolumn{2}{l}{\textit{Generalized Formulation (Appendix)}} \\
    $N$ & Number of conditional contexts (constraints). \\
    $\mathcal{C} = \{C_k\}$ & The set of $N$ distinct context conditions ($k=1 \dots N$). \\
    $\mathcal{X}_t = \{\mathbf{x}_t^{(k)}\}$ & \textbf{System State:} Set of all $N+1$ parallel latent states at time $t$. \\
    $\mathbf{x}_t^{(k)}$ & The latent state of the $k$-th branch at timestep $t$. \\
    $k=N+1$ & Index denoting the unconditional (prior) branch. \\
    $\mathcal{M}$ & Binary mask partitioning the state into Subject ($1$) and Context ($0$). \\
    $S, \mathbf{x}[\mathcal{M}]$ & General notation for the Subject region (equivalent to $B$ in main text). \\
    $t, T$ & Current diffusion timestep and total diffusion steps. \\
    $\hat{\mathbf{x}}_0^{(k)}$ & Predicted clean data estimate at time $t$ for branch $k$. \\
    \midrule
    \multicolumn{2}{l}{\textit{Scheduling \& Annealing Parameters}} \\
    $\mathcal{S}_{sync}$ & \textbf{Synchronization Scope:} The set of timesteps where consensus is potentially active. \\
    $\mathcal{S}_{heat}$ & \textbf{Annealing Scope:} Heating schedule mapping $t \mapsto (j_t, K_t)$. \\
    $T_{start}, T_{end}$ & The start and end timestamps defining the active window. \\
    $J_{heat}$ & \textbf{Jump Size:} Number of steps to jump back during the heating phase. \\
    $K$ & \textbf{Resampling Iterations:} Number of heating-cooling cycles per timestep. \\
    \midrule
    \multicolumn{2}{l}{\textit{Aggregation Parameters}} \\
    $B_{canon}$ & The computed canonical consensus subject. \\
    $\mu_{cond}$ & Average prediction of the subject from all $N$ conditional branches. \\
    $\lambda$ & Guidance scale controlling aggregation ($\lambda=0$: Mean, $\lambda \approx N$: PoE). \\
    \bottomrule
    \end{tabularx}
\end{table}
\section{Additional Experimental Results}
\label{app:exp_results}

This appendix provides supplementary experimental results for both application domains. We present the material in the same order as the main text: flow field reconstruction followed by co-antibody generation.

\subsection{Flow Field Reconstruction}
\label{app:flow_results}

The complete experimental results for flow field reconstruction are presented in the main text. Specifically, Table~\ref{tab:flow_field} reports the quantitative comparison on standard reconstruction tasks, and Table~\ref{tab:flow_zero_shot} presents zero-shot generalization results. Detailed analysis of these results can be found in Section~\ref{subsec:flow_results}.

\subsection{Co-antibody Generation}
\label{app:antibody_results}

\subsubsection{Full Ablation Study}
\label{app:ablation_results}

In this section, we provide the comprehensive results of our thermodynamic scheduling ablation studies. While the main text highlights key trends and the best configurations, Table~\ref{tab:ablation_full} presents the detailed performance across all 24 evaluated configurations. These experiments validate the robustness of our framework across varying thermodynamic regimes and target clusters.

\paragraph{Data Selection Criteria (Split-A).}
The ablation targets were curated as a specific subset, denoted as \textbf{Split-A}, focusing on computational efficiency. We selected the \textbf{10 smallest clusters} (as detailed in Appendix~\ref{app:dataset_details}) where the total protein complex size allows for rapid experimental cycles during extensive thermodynamic grid searches. This subset ensures that the most computationally intensive part of our evaluation—exploring the transition between energy injection and manifold relaxation—is performed on targets that provide high-resolution feedback with minimal overhead.

\paragraph{Design Space of Thermodynamic Schedules.}
\label{app:design_space}
We identify three primary degrees of freedom in our framework's configuration:
\begin{itemize}
    \item \textbf{Cooling-Heating Cycle:} We compare two schedule archetypes: (i) \textit{Periodic Resampling}, which alternates between cooling $J_{heat}$ steps and heating $J_{heat}$ steps for $K$ iterations; and (ii) \textit{Continuous Annealing}, which cools the system to the final state once before reheating back to a noise level $t_{heat}$ and initiating a second cooling pass.
    \item \textbf{Synchronization Policy:} We evaluate when to enforce consensus. In a multi-pass schedule, we compare the \textbf{``First-Visit Only''} policy (synchronizing only during the initial cooling pass) against the \textbf{``Consistent ''} policy (synchronizing during every visit to a timestep).
    \item \textbf{Heating Intensity:} For a fixed schedule, we adjust the jump size $J_{heat}$, iterations $K$, and the maximum noise height (Heat Height) to control the exploration-exploitation balance.
\end{itemize}

\paragraph{Detailed Analysis of Thermodynamic Parameters.}
Our ablation study systematically explores three key dimensions of the annealing process:

\textit{Exploration Depth ($K$):} We investigate the impact of heating iterations with a fixed jump size $J_{heat}=3$. Employing a grid search over $K \in \{1, 2, 3, 4\}$, we observe that performance peaks at $K=2$. While moderate reheating facilitates refinement, increasing the depth to $K=3$ or $K=4$ leads to performance degradation. This suggests that excessive energy injection may destabilize the trajectory, pushing the system too far from the consensus manifold established during the initial cooling pass.

\textit{Noise Scale (Heat Height):} The Heat Height parameter scales the magnitude of noise injection during the reheating phase. Our resampling mechanism draws inspiration from RePaint~\cite{lugmayr2022repaint} and annealed Langevin dynamics~\cite{song2019generative, song2020score}, where high-temperature resets are theoretically required to ensure sufficient mixing to traverse energy barriers. Empirically, we find that full energy injection (Height 1.00) yields optimal results, whereas reduced heights are less effective at escaping the local minima characteristic of conflicting multi-target constraints.

\textit{Synchronization Windows:} We evaluate distinct temporal strategies for applying the consensus operator, validating hypotheses from prior work on dynamic guidance~\cite{yu2023freedom, bansal2024universal} and collaborative diffusion~\cite{huang2023collaborative}. The results reveal a clear trade-off: \textit{No Synchronization} fails to achieve alignment, while \textit{Full Step Synchronization} over-constrains the generative process, restricting necessary structural relaxation. Consequently, the \textit{``First-Visit Only''} policy emerges as the optimal strategy, applying constraints strategically to establish global consensus without impeding local refinement.

\begin{table*}[t]
\centering
\caption{Full ablation results on Split-A Weighted Fusion (Mean aggregation). This table consolidates all thermodynamic schedule grid searches.}
\label{tab:ablation_full}
\begin{tabular}{lcccccc}
\toprule
\multirow{2}{*}{\textbf{Configuration}} & \multicolumn{3}{c}{\textbf{RMSD} (\AA) (\(\downarrow\))} & \multicolumn{3}{c}{\textbf{design-ipTM} (\(\uparrow\))} \\
\cmidrule(lr){2-4} \cmidrule(lr){5-7}
 & Ag. A & Ag. C & Avg. & Ag. A & Ag. C & Avg. \\
\midrule
\multicolumn{7}{l}{\small\textbf{\textit{Exploration Depth (Iterations $K$)}}} \\
$J_{heat}=2, K=1$ & 3.80 & 3.81 & 3.80 & 0.19 & 0.19 & 0.19 \\
$J_{heat}=2, K=2$ & 3.87 & 3.89 & 3.88 & 0.18 & 0.18 & 0.18 \\
$J_{heat}=2, K=3$ & 3.89 & 3.87 & 3.88 & 0.19 & 0.19 & 0.19 \\
$J_{heat}=2, K=4$ & 3.91 & 4.00 & 3.96 & 0.18 & 0.18 & 0.18 \\
$J_{heat}=3, K=1$ & 3.76 & 3.77 & 3.76 & 0.19 & 0.19 & 0.19 \\
$J_{heat}=3, K=2$  & \textbf{3.75} & 3.77 & \textbf{3.76} & \textbf{0.19} & \textbf{0.19} & \textbf{0.19} \\
$J_{heat}=3, K=3$ & 3.88 & 3.95 & 3.92 & 0.19 & 0.19 & 0.19 \\
$J_{heat}=3, K=4$ & 3.91 & 3.86 & 3.89 & 0.18 & 0.18 & 0.18 \\
$J_{heat}=4, K=1$ & 3.80 & 3.80 & 3.80 & 0.19 & 0.19 & 0.19 \\
$J_{heat}=4, K=2$ & 3.99 & 3.95 & 3.97 & 0.18 & 0.19 & 0.19 \\
$J_{heat}=4, K=3$ & 3.95 & 3.89 & 3.92 & 0.18 & 0.18 & 0.18 \\
$J_{heat}=4, K=4$ & 4.03 & 4.01 & 4.02 & 0.19 & 0.18 & 0.19 \\
$J_{heat}=5, K=1$ & 3.82 & \textbf{3.74} & 3.78 & 0.19 & 0.19 & 0.19 \\
$J_{heat}=5, K=2$ & 3.89 & 3.91 & 3.90 & 0.19 & 0.19 & 0.19 \\
$J_{heat}=5, K=3$ & 4.04 & 4.04 & 4.04 & 0.18 & 0.18 & 0.18 \\
$J_{heat}=5, K=4$ & 3.96 & 4.02 & 3.99 & 0.19 & 0.19 & 0.19 \\
\midrule
\multicolumn{7}{l}{\small\textbf{\textit{Noise Scale (Heat Height)}}} \\
Height 1.00  & \textbf{3.75} & 3.77 & \textbf{3.76} & \textbf{0.19} & 0.19 & 0.19 \\
Height 0.75 & 3.88 & \textbf{3.72} & 3.80 & 0.19 & 0.19 & 0.19 \\
Height 0.50 & 3.83 & 3.78 & 3.80 & 0.19 & 0.19 & 0.19 \\
Height 0.25 & 3.83 & 3.81 & 3.82 & 0.19 & \textbf{0.20} & \textbf{0.20} \\
\midrule
\multicolumn{7}{l}{\small\textbf{\textit{Synchronization Windows}}} \\
Windowed Synchronization & \textbf{3.77} & \textbf{3.80} & \textbf{3.79} & 0.18 & 0.19 & 0.19 \\
Full Step Synchronization & 3.87 & 3.90 & 3.89 & \textbf{0.19} & \textbf{0.19} & \textbf{0.19} \\
No Synchronization & 4.14 & 4.01 & 4.07 & 0.18 & 0.20 & 0.19 \\
\bottomrule
\end{tabular}
\end{table*}

\paragraph{Comparison of Consensus Strategies.}
The results, presented in Table~\ref{tab:ablation_full_dc}, align with our findings in the main text regarding the superior stability of weighted fusion. We observe that \textbf{Joint Overlap Factorization} (DiffCollage-style) exhibits higher sensitivity to synchronization frequency compared to \textbf{Weighted Fusion} (Mean aggregation). Notably, while increasing synchronization steps generally benefits Weighted Fusion (Table~\ref{tab:ablation_full}), it can degrade performance for Overlap Factorization. This suggests that the gradient-based updates in the overlap formulation may conflict with the delicate structural constraints of the antibody-antigen interface, introducing high-frequency artifacts. In contrast, the geometric averaging of Weighted Fusion provides a more robust projection onto the valid consensus manifold, making it a safer default choice across varying annealing schedules.

\begin{table*}[ht]
\centering
\caption{Full ablation results on Split-A using Joint Overlap Factorization (DiffCollage-style) aggregation. Note that DiffCollage consistently underperforms compared to Mean aggregation (Table~\ref{tab:ablation_full}) in this high-dimensional constraints setting.}
\label{tab:ablation_full_dc}
\begin{tabular}{lcccccc}
\toprule
\multirow{2}{*}{\textbf{Configuration}} & \multicolumn{3}{c}{\textbf{RMSD} (\AA) (\(\downarrow\))} & \multicolumn{3}{c}{\textbf{design-ipTM} (\(\uparrow\))} \\
\cmidrule(lr){2-4} \cmidrule(lr){5-7}
 & Ag. A & Ag. C & Avg. & Ag. A & Ag. C & Avg. \\
\midrule
\multicolumn{7}{l}{\small\textbf{\textit{Exploration Depth (Iterations $K$)}}} \\
$J_{heat}=2, K=1$ & 4.55 & 4.47 & 4.51 & \textbf{0.20} & \textbf{0.20} & \textbf{0.20} \\
$J_{heat}=2, K=2$ & 4.60 & 4.54 & 4.57 & 0.20 & 0.20 & 0.20 \\
$J_{heat}=2, K=3$ & 4.53 & 4.58 & 4.56 & 0.19 & 0.19 & 0.19 \\
$J_{heat}=2, K=4$ & 4.46 & 4.44 & 4.45 & 0.18 & 0.19 & 0.19 \\
$J_{heat}=3, K=1$ & 4.54 & 4.51 & 4.53 & 0.20 & 0.20 & 0.20 \\
$J_{heat}=3, K=2$ & 4.46 & 4.40 & 4.43 & 0.19 & 0.19 & 0.19 \\
$J_{heat}=3, K=3$ & 4.49 & 4.48 & 4.49 & 0.19 & 0.19 & 0.19 \\
$J_{heat}=3, K=4$ & 4.41 & 4.33 & 4.37 & 0.19 & 0.19 & 0.19 \\
$J_{heat}=4, K=1$ & 4.56 & 4.53 & 4.55 & 0.20 & 0.20 & 0.20 \\
$J_{heat}=4, K=2$ & 4.44 & 4.38 & 4.41 & 0.19 & 0.19 & 0.19 \\
$J_{heat}=4, K=3$ & 4.43 & 4.38 & 4.40 & 0.19 & 0.19 & 0.19 \\
$J_{heat}=4, K=4$ & \textbf{4.27} & 4.32 & \textbf{4.30} & 0.18 & 0.18 & 0.18 \\
$J_{heat}=5, K=1$ & 4.57 & 4.55 & 4.56 & 0.20 & 0.20 & 0.20 \\
$J_{heat}=5, K=2$ & 4.39 & 4.42 & 4.40 & 0.19 & 0.19 & 0.19 \\
$J_{heat}=5, K=3$ & 4.30 & 4.39 & 4.35 & 0.19 & 0.19 & 0.19 \\
$J_{heat}=5, K=4$ & 4.30 & \textbf{4.29} & 4.30 & 0.18 & 0.18 & 0.18 \\
\midrule
\multicolumn{7}{l}{\small\textbf{\textit{Noise Scale (Heat Height)}}} \\
Height 1.00  & \textbf{4.27} & 4.32 & \textbf{4.30} & 0.18 & 0.18 & 0.18 \\
Height 0.75 & 4.45 & 4.59 & 4.52 & \textbf{0.20} & \textbf{0.20} & \textbf{0.20} \\
Height 0.50 & 4.54 & 4.56 & 4.55 & 0.20 & 0.20 & 0.20 \\
Height 0.25 & 4.46 & 4.56 & 4.51 & 0.20 & 0.20 & 0.20 \\
\midrule
\multicolumn{7}{l}{\small\textbf{\textit{Synchronization Windows}}} \\
Windowed Synchronization & \textbf{4.39} & \textbf{4.35} & \textbf{4.37} & 0.18 & 0.20 & 0.19 \\
Full Step Synchronization & 4.40 & 4.54 & 4.47 & 0.19 & 0.19 & 0.19 \\
No Synchronization & 4.92 & 4.83 & 4.88 & \textbf{0.20} & \textbf{0.20} & \textbf{0.20} \\
\bottomrule
\end{tabular}
\end{table*}

\subsubsection{Per-Target Performance}
\label{app:per_target_results}

Table~\ref{tab:antibody_results_per_target} provides a granular breakdown of performance across all 10 antigen pairs in the Split-B benchmark. This detailed view reveals the structural heterogeneity of the test cases and highlights the robustness of our proposed method.

The \textit{Single-target} baseline exhibits significant variance, performing well on simpler pairs (e.g., \texttt{2r4r/2r4s}) but struggling on more complex cases (e.g., \texttt{6j5d/6j5f}, where RMSD degrades to 3.95~\AA). In contrast, \textbf{ACG/w} maintains consistent sub-3.5~\AA\ precision across nearly all cases.
Notably, in challenging scenarios such as \texttt{9ptm/9me5} and \texttt{6j5d/6j5f}, naive strategies like \textit{Consistent } suffer from catastrophic structural distortions (RMSD $>4.9$~\AA), likely due to over-constraining the trajectory. ACG/w effectively resolves these conflicts, achieving valid conformations with RMSDs of 2.83~\AA\ and 3.26~\AA, respectively. Even in cases where the baseline is strong (e.g., \texttt{2r4r/2r4s}), ACG/w remains competitive (3.01~\AA), demonstrating that it does not degrade performance on easier targets while offering substantial gains on harder ones.

\paragraph{Impact of Weighted Fusion.}
Consistent with our observations in the ablation study, the weighted fusion operator (denoted by \textit{/w}) acts as a universal stabilizer across different generation strategies. As shown in the table, \textit{Greedy/w} and \textit{Consistent /w} consistently outperform their unweighted counterparts (Standard Overlap). However, simply adding weighted fusion is insufficient; for instance, \textit{Consistent /w} still lags behind \textit{ACG/w} in 8 out of 10 cases. This confirms that the "First-Visit" annealing schedule—not just the fusion operator—is the primary driver of ACG's superior performance.
\clearpage
\begin{longtable}{llrrrrrr}
\caption{\textbf{Detailed per-target performance breakdown on the high-fidelity Split-B benchmark.} This table expands upon the aggregate results in Table~\ref{tab:antibody_results}, providing granular metrics for each of the 10 distinct antigen pairs. We compare the Independent Oracle (Single-target) against various joint optimization configurations. \textbf{ACG/w} consistently achieves the best balance of structural precision (RMSD) and binding confidence (design-ipTM) across diverse target topologies. Best performance is bolded.}
\label{tab:antibody_results_per_target} \\
\toprule
\multirow{2}{*}{\textbf{Target Pair}} & \multirow{2}{*}{\textbf{Configuration}} & \multicolumn{3}{c}{\textbf{RMSD} (\AA) (\(\downarrow\))} & \multicolumn{3}{c}{\textbf{design-ipTM} (\(\uparrow\))} \\
\cmidrule(lr){3-5} \cmidrule(lr){6-8}
 & & \multicolumn{1}{c}{Ag. A} & \multicolumn{1}{c}{Ag. C} & \multicolumn{1}{c}{Avg.} & \multicolumn{1}{c}{Ag. A} & \multicolumn{1}{c}{Ag. C} & \multicolumn{1}{c}{Avg.} \\
\midrule
\endfirsthead

\multicolumn{8}{c}{\tablename\ \thetable{} -- continued from previous page} \\
\toprule
\multirow{2}{*}{\textbf{Target Pair}} & \multirow{2}{*}{\textbf{Configuration}} & \multicolumn{3}{c}{\textbf{RMSD} (\AA) (\(\downarrow\))} & \multicolumn{3}{c}{\textbf{design-ipTM} (\(\uparrow\))} \\
\cmidrule(lr){3-5} \cmidrule(lr){6-8}
 & & \multicolumn{1}{c}{Ag. A} & \multicolumn{1}{c}{Ag. C} & \multicolumn{1}{c}{Avg.} & \multicolumn{1}{c}{Ag. A} & \multicolumn{1}{c}{Ag. C} & \multicolumn{1}{c}{Avg.} \\
\midrule
\endhead

\midrule
\multicolumn{8}{r}{Continued on next page} \\
\endfoot

\bottomrule
\endlastfoot

\multirow{7}{*}{\texttt{4jre/4jr9}} & Single-target & 2.80 & 2.97 & 2.89 & \textbf{0.47} & 0.31 & 0.39 \\
 & Greedy & 3.24 & 3.21 & 3.23 & 0.35 & 0.36 & 0.36 \\
 & Greedy/w & 2.72 & 2.92 & 2.82 & 0.34 & 0.38 & 0.36 \\
 & Consistent  & 3.30 & 3.32 & 3.31 & 0.39 & 0.41 & 0.40 \\
 & Consistent /w & 2.77 & \textbf{2.59} & 2.68 & 0.43 & 0.41 & 0.42 \\
 & ACG & 3.31 & 3.19 & 3.25 & 0.43 & 0.45 & 0.44 \\
 \rowcolor{red!8} \cellcolor{white}  & ACG/w & \textbf{2.51} & 2.66 & \textbf{2.59} & 0.42 & \textbf{0.45} & \textbf{0.44} \\
\midrule
\multirow{7}{*}{\texttt{7zf9/7bei}} & Single-target & 3.24 & \textbf{2.75} & 3.00 & \textbf{0.35} & 0.22 & 0.29 \\
 & Greedy & 3.43 & 3.73 & 3.58 & 0.33 & 0.29 & 0.31 \\
 & Greedy/w & 3.06 & 3.00 & 3.03 & 0.31 & 0.26 & 0.29 \\
 & Consistent  & 3.28 & 3.30 & 3.29 & 0.34 & 0.30 & 0.32 \\
 & Consistent /w & 2.81 & 2.87 & 2.84 & 0.32 & 0.27 & 0.30 \\
 & ACG & 3.41 & 3.11 & 3.26 & 0.34 & \textbf{0.30} & \textbf{0.32} \\
 \rowcolor{red!8} \cellcolor{white}  & ACG/w & \textbf{2.62} & 2.84 & \textbf{2.73} & 0.30 & 0.29 & 0.30 \\
\midrule
\multirow{7}{*}{\texttt{5tru/7elx}} & Single-target & 3.01 & 3.22 & 3.12 & 0.23 & 0.23 & 0.23 \\
 & Greedy & 3.68 & 3.60 & 3.64 & 0.22 & 0.22 & 0.22 \\
 & Greedy/w & 3.27 & 3.16 & 3.22 & 0.22 & 0.21 & 0.22 \\
 & Consistent  & 3.67 & 3.57 & 3.62 & 0.22 & 0.22 & 0.22 \\
 & Consistent /w & 3.08 & \textbf{2.86} & 2.97 & 0.23 & 0.23 & 0.23 \\
 & ACG & 3.33 & 3.47 & 3.40 & 0.23 & 0.21 & 0.22 \\
 \rowcolor{red!8} \cellcolor{white}  & ACG/w & \textbf{2.85} & 2.91 & \textbf{2.88} & \textbf{0.24} & \textbf{0.23} & \textbf{0.24} \\
\midrule
\multirow{7}{*}{\texttt{4yxk/4h88}} & Single-target & 3.09 & 3.01 & 3.05 & 0.24 & 0.28 & 0.26 \\
 & Greedy & 3.18 & 3.27 & 3.23 & 0.31 & 0.32 & 0.32 \\
 & Greedy/w & 2.91 & 3.00 & 2.96 & 0.27 & 0.29 & 0.28 \\
 & Consistent  & 3.46 & 3.49 & 3.48 & \textbf{0.32} & \textbf{0.32} & \textbf{0.32} \\
 & Consistent /w & 2.85 & \textbf{2.80} & 2.83 & 0.26 & 0.26 & 0.26 \\
 & ACG & 3.34 & 3.21 & 3.28 & 0.27 & 0.29 & 0.28 \\
 \rowcolor{red!8} \cellcolor{white}  & ACG/w & \textbf{2.79} & 2.86 & \textbf{2.83} & 0.24 & 0.26 & 0.25 \\
\midrule
\multirow{7}{*}{\texttt{2r4r/2r4s}} & Single-target & \textbf{2.76} & \textbf{3.03} & \textbf{2.89} & \textbf{0.61} & 0.31 & \textbf{0.46} \\
 & Greedy & 3.66 & 3.58 & 3.62 & 0.46 & \textbf{0.43} & 0.45 \\
 & Greedy/w & 3.29 & 3.25 & 3.27 & 0.47 & 0.37 & 0.42 \\
 & Consistent  & 4.38 & 4.22 & 4.30 & 0.42 & 0.39 & 0.41 \\
 & Consistent /w & 3.03 & 3.22 & 3.13 & 0.36 & 0.35 & 0.36 \\
 & ACG & 3.63 & 3.78 & 3.71 & 0.47 & 0.38 & 0.43 \\
 \rowcolor{red!8} \cellcolor{white}  & ACG/w & 2.94 & 3.09 & 3.01 & 0.40 & 0.36 & 0.38 \\
\midrule
\multirow{7}{*}{\texttt{9ptm/9me5}} & Single-target & 3.01 & 3.32 & 3.17 & 0.38 & 0.31 & 0.35 \\
 & Greedy & 3.97 & 3.84 & 3.91 & 0.39 & 0.37 & 0.38 \\
 & Greedy/w & 3.25 & 3.47 & 3.36 & 0.36 & 0.33 & 0.35 \\
 & Consistent  & 5.30 & 5.31 & 5.31 & \textbf{0.40} & \textbf{0.39} & \textbf{0.40} \\
 & Consistent /w & 3.08 & 2.97 & 3.03 & 0.35 & 0.36 & 0.36 \\
 & ACG & 3.96 & 4.07 & 4.02 & 0.37 & 0.35 & 0.36 \\
 \rowcolor{red!8} \cellcolor{white}  & ACG/w & \textbf{2.77} & \textbf{2.89} & \textbf{2.83} & 0.35 & 0.37 & 0.36 \\
\midrule
\multirow{7}{*}{\texttt{7lkh/7lkf}} & Single-target & 3.79 & 3.91 & 3.85 & \textbf{0.34} & 0.23 & \textbf{0.29} \\
 & Greedy & 4.33 & 4.18 & 4.26 & 0.29 & 0.24 & 0.27 \\
 & Greedy/w & 3.69 & 3.64 & 3.67 & 0.31 & 0.21 & 0.26 \\
 & Consistent  & 5.13 & 5.14 & 5.14 & 0.32 & 0.23 & 0.28 \\
 & Consistent /w & 3.42 & 3.44 & 3.43 & 0.26 & 0.21 & 0.24 \\
 & ACG & 4.02 & 4.08 & 4.05 & 0.31 & \textbf{0.25} & 0.28 \\
 \rowcolor{red!8} \cellcolor{white}  & ACG/w & \textbf{3.13} & \textbf{3.14} & \textbf{3.14} & 0.29 & 0.24 & 0.27 \\
\midrule
\multirow{7}{*}{\texttt{7x1m/8hwt}} & Single-target & 3.40 & 3.35 & 3.38 & 0.35 & 0.33 & 0.34 \\
 & Greedy & 3.73 & 3.59 & 3.66 & 0.38 & 0.37 & 0.38 \\
 & Greedy/w & 3.42 & 3.36 & 3.39 & 0.36 & 0.35 & 0.36 \\
 & Consistent  & 3.79 & 3.75 & 3.77 & \textbf{0.41} & 0.38 & \textbf{0.40} \\
 & Consistent /w & 3.23 & 3.49 & 3.36 & 0.37 & 0.37 & 0.37 \\
 & ACG & 3.87 & 3.59 & 3.73 & 0.38 & 0.38 & 0.38 \\
 \rowcolor{red!8} \cellcolor{white}  & ACG/w & \textbf{3.01} & \textbf{3.18} & \textbf{3.09} & 0.39 & \textbf{0.38} & 0.39 \\
\midrule
\multirow{7}{*}{\texttt{6j5d/6j5f}} & Single-target & 3.59 & 4.30 & 3.95 & 0.32 & 0.40 & 0.36 \\
 & Greedy & 5.13 & 5.11 & 5.12 & 0.32 & 0.39 & 0.36 \\
 & Greedy/w & 3.38 & 3.72 & 3.55 & 0.28 & 0.35 & 0.32 \\
 & Consistent  & 4.97 & 4.93 & 4.95 & \textbf{0.33} & \textbf{0.40} & \textbf{0.37} \\
 & Consistent /w & 3.45 & 3.57 & 3.51 & 0.27 & 0.34 & 0.31 \\
 & ACG & 4.46 & 4.20 & 4.33 & 0.31 & 0.39 & 0.35 \\
 \rowcolor{red!8} \cellcolor{white}  & ACG/w & \textbf{3.38} & \textbf{3.15} & \textbf{3.26} & 0.30 & 0.35 & 0.32 \\
\midrule
\multirow{7}{*}{\texttt{5tr1/5tqq}} & Single-target & \textbf{3.40} & 3.96 & 3.68 & 0.38 & 0.32 & 0.35 \\
 & Greedy & 4.33 & 4.54 & 4.44 & 0.35 & 0.34 & 0.35 \\
 & Greedy/w & 3.68 & 3.81 & 3.75 & \textbf{0.41} & \textbf{0.36} & \textbf{0.39} \\
 & Consistent  & 4.90 & 5.02 & 4.96 & 0.34 & 0.33 & 0.34 \\
 & Consistent /w & 3.74 & \textbf{3.44} & 3.59 & 0.34 & 0.35 & 0.35 \\
 & ACG & 4.21 & 4.31 & 4.26 & 0.36 & 0.35 & 0.36 \\
 \rowcolor{red!8} \cellcolor{white}  & ACG/w & 3.60 & 3.47 & \textbf{3.54} & 0.36 & 0.33 & 0.35 \\
\end{longtable}

\end{document}